\documentclass[11pt,a4paper]{article}
\usepackage[hyperref]{acl2020}
\usepackage{times}
\usepackage{latexsym}

\usepackage{microtype}

\usepackage{mathtools}
\usepackage{graphicx}
\usepackage{amsmath}
\usepackage{adjustbox}
\usepackage{caption}
\usepackage{booktabs}
\usepackage{makecell}
\usepackage{multirow}
\usepackage{amsmath,environ}
\usepackage{tabularx}
\usepackage{subcaption}
\usepackage{graphicx}
\usepackage{comment}
\usepackage{amssymb}
\usepackage{pifont}
\usepackage{times}
\usepackage{latexsym}
\usepackage{url}

\usepackage{hyperref}
\usepackage{url}
\DeclareMathOperator{\E}{\mathbb{E}}
\aclfinalcopy 
\def\aclpaperid{2071} 

\title{Learning Interpretable and Discrete Representations \\ with Adversarial Training for Unsupervised Text Classification }

\author{Yau-Shian Wang\quad Hung-Yi Lee\quad Yun-Nung Chen \\National Taiwan University, Taipei, Taiwan\\
{\tt king6101@gmail.com\quad hungyilee@ntu.edu.tw\quad y.v.chen@ieee.org}\\}

\date{}

\begin{document}
\maketitle
\begin{abstract}
Learning continuous representations from unlabeled textual data has been increasingly studied for benefiting semi-supervised learning.
Although it is relatively easier to interpret discrete representations, due to the difficulty of training, learning discrete representations for unlabeled textual data has not been widely explored.
This work proposes TIGAN that learns to encode texts into two disentangled representations, including a discrete code and a continuous noise, where the discrete code represents interpretable topics, and the noise controls the variance within the topics.
The discrete code learned by TIGAN can be used for unsupervised text classification.
Compared to other unsupervised baselines, the proposed TIGAN achieves superior performance on six different corpora.
Also, the performance is on par with a recently proposed weakly-supervised text classification method.
The extracted topical words for representing latent topics show that TIGAN learns coherent and highly interpretable topics.
\end{abstract}

\section{Introduction}
In natural language processing (NLP), learning meaningful representations from large amounts of unlabeled texts is a core problem for unsupervised and semi-supervised language understanding.
While learning continuous text representations has been widely studied~\cite{skip-thought,tough-to-beat-embedding,efficient-sentence-representation,n-Gram, elmo, BERT}, learning discrete representations has been explored by fewer works~\cite{discrete-topic, Discrete-Sentence-Representation}.
Nevertheless, learning discrete representations is still important, as discrete representations are easier to interpret~\cite{Discrete-Sentence-Representation} and can benefit unsupervised learning~\cite{VQ-VAE}. 

Auto-encoders and their variants are widely used to learn latent representations, but how to learn meaningful \emph{discrete} latent representations remains unsolved. %
The reason is that using discrete variables as latent representations in auto-encoders hinders the gradient from backpropagating from the decoder to the encoder. %
To take non-differentiable discrete variables as latent representations in an auto-encoder, some special methods, such as Gumbel-Softmax~\cite{Gumbel-Softmax} or vector quantization~\cite{VQ-VAE}, are applied to enable training.
Another direction of learning useful discrete representations is to maximize the mutual information between data and discrete variables such as infoGAN~\cite{Info-GAN} or IMSAT~\cite{IMSAT}. 

In this work, we propose Textual InfoGAN (TIGAN), which is mainly built upon InfoGAN but with several useful extensions to make it suitable for textual data. 
In this model, texts are generated from two disentangled representations, including a discrete code $c$ and a continuous noise $z$, where each dimension of $c$ represents a topic (or a category), and $z$ controls the variance within a topic. 
For example, if the model learns that one dimension of $c$ represents a topic about ``sport'', different $z$ represents different sport types like basketball or baseball, or different teams.
Given a new text, the model can efficiently infer the discrete topic code $c$. 
As we alternatively optimize the InfoGAN objective and auto-encode objective function, the model can be considered as the integration of the infoGAN and auto-encoder based approaches. 

To evaluate the interpretability of discovered discrete topics, we evaluate our model on unsupervised text classification.
Better performance on unsupervised text classification implies that the discovered topics directly match the human-annotated categories, and thus humans can intuitively understand what they represent, such as the category of news or the type of questions. 
Compared to other possible unsupervised text classification methods, such as unsupervised sentence representation methods or topic models, TIGAN achieves superior performance.
Also, the performance of our unsupervised method is on par with a recent weakly-supervised method ~\citep{FAC}, which required keywords of each category annotated by human experts.

To interpret discovered topics, we extract a few topical words from each topic to represent the topic.
The extracted topical words are evaluated by quantitative and qualitative analysis.
Compared to baseline methods, TIGAN is capable of extracting more coherent topical words.
Also, it is possible to generate various texts conditioned on topics discovered by TIGAN in an unsupervised manner, which gives us a way to interpret the discovered topics further.
The generated texts demonstrate that TIGAN successfully learns disentangled representations.

\section{Related Work}
Here we review the approaches of unsupervised discrete representation learning.
\paragraph{Auto-encoder}
Our work is closely related to other unsupervised discrete representation learning methods.
VQ-VAE~\cite{VQ-VAE} encodes the data into a discrete one-hot code by drawing the index with embedding closest to the data representation.
They utilize vector quantization (VQ) to approximate the gradient from the decoder to the encoder.
DI-VAE~\cite{Discrete-Sentence-Representation} uses Batch Prior Regularization (BPR) to approximate the KL-divergence between discrete variables, and learn discrete representations by Gumbel-Softmax and minimizing the KL-divergence between discrete posterior and a discrete prior.

In this work, because we set the dimension of discrete code to a small number, we have to model the variance within the code.
Otherwise, the text can not be successfully generated.
Therefore, we search for other discrete representation learning models.

\paragraph{Maximizing Mutual Information} 
To learn useful latent representation, IMSAT~\cite{IMSAT} maximizes the mutual information between data and the encoded discrete representation.
They also propose an objective function to make the discrete representation invariant to the data augmentation.
On the other hand, InfoGAN has shown impressive performance for learning disentangled representations of images in an unsupervised manner~\cite{Info-GAN}.
The original GAN generates images from a continuous noise $z$, while each dimension of the noise does not contain disentangled features of generated images. 
To learn semantically meaningful representations, InfoGAN maximizes the mutual information between input code $c$ and the generated output $G(z, c)$. 
However, maximizing the mutual information is intractable because it requires the access of $P(c\mid G(z, c))$.
Based on variational information maximization, \cite{Info-GAN} used an auxiliary function $Q$ to approximate $P(c\mid G(z, c))$.
The auxiliary function $Q$ can be a neural network and jointly optimized with $G$.

\section{TIGAN}

\begin{figure}[t]
  \centering
    \includegraphics[width=\linewidth]{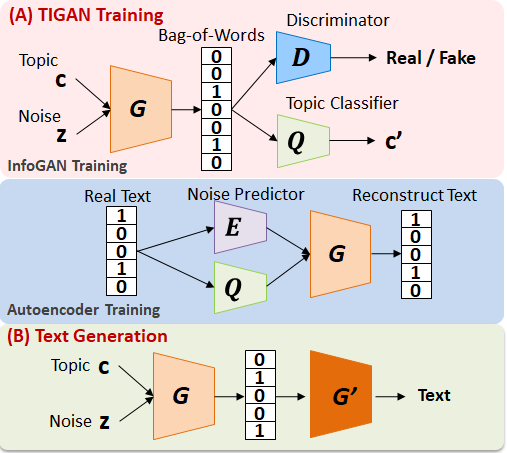}
      \caption{Illustration of the proposed model. \textbf{(A)} In the upper part (InfoGAN training), the bag-of-words generator $G$ takes discrete topic code  $c$ and continuous noise vector $z$ as input and generates bag-of-words. 
      The $D$ discriminates its input bag-of-words is from $G$ or human-written text. 
      The topic classifier $Q$ predicts the latent topic $c$ from its input bag-of-words. 
      In the lower part, an auto-encoder is learned from bag-of-words of human-written text. 
      The noise predictor $E$ which predicts the noise $z$ from input bag-of-words and the topic classifier $Q$ together form the encoder, while $G$ is the decoder. 
      \textbf{(B)} To interpret what is learned by the model, an LSTM text generator $G^\prime$ is trained to generate text based on bag-of-words. } 
  \label{fig:topic_gan}
\end{figure}

TIGAN considers that text is generated from a discrete code $c$ and a continuous noise $z$. 
Each dimension of discrete code $c$ represents a topic (or a category), and the continuous noise $z$ controls the variance within the topic. 
As sequential textual data is too complex to discover meaningful topics, we focus on learning topics from bag-of-words textual data.
Besides, we propose some extra extensions of infoGAN including:
(1) categorical loss clipping, 
(2) combining infoGAN with auto-encoder,
(3) using WGAN-gp~\cite{WGAN-gp}, 
and (4) using pretrained word embeddings to regularize the discovered topics.
The above extensions greatly improve performance.

\subsection{Model} \label{sec:model}
The Figure~\ref{fig:topic_gan}~(A) illustrates   our model, where there are a generator $G$, a discriminator $D$, a topic classifier $Q$, and a noise predictor $E$:
\begin{itemize}
    \item Generator $G$\\
    It takes a discrete topic code $c$ and a continuous noise $z$ as the input, and manages to generate bag-of-words that are indistinguishable from the bag-of-words of real texts, while captures the topical information of input $c$. 
      Here the output of the generator, $G(c,z)$,  is a vocabulary size vector.
        The output layer of $G$ is $sigmoid$ function, so the value of each dimension is between $0$ and $1$.
        $G(c,z)$ can be interpreted as a bag-of-words vector, where each dimension indicates whether a single word exists in the text. 
    The $G(c,z)$ is directly fed to the discriminator, topic classifier, and noise predictor without sampling.
    \item Discriminator $D$\\
    It takes a bag-of-word vector as its input and outputs a scalar, which distinguishes whether the input is generated by the generator $G$  or from human-written texts.
  Here the bag-of-words vector $x$ from human-written texts would be assigned a higher score, or $D(x)$ would be larger.
    On the other hand, the bag-of-words vector $G(c,z)$ generated by $G$ would be assigned lower score by $D$, or  $D(G(c,z))$ would be smaller.
    $D$ is trained to minimize the following loss function $\mathcal{L}_D$
\begin{equation}  \label{eq:L_D}
\begin{split}
\mathcal{L}_D = &  -\E_{x \sim P_{data}}[\log(D(x))] \\ & -\E_{z \sim P_{z},c \sim P_{c}}[\log(1-D(G(z, c)))],
\end{split}
\end{equation}
    where $P_{data}$ is the real data distribution, $P_{z}$ is the noise distribution, and $P_{c}$ is the topic code distribution.     
    The distributions  $P_{c}$ and $P_{z}$ are determined by the developers based on the prior knowledge about the corpus.
    In the following experiments, $P_{z}$ is a normal distribution, while $P_{c}$ is a uniform distribution that generates one-hot vectors in which each dimension has an equal probability of being set to $1$.
    In general, we find using the normal distribution as $P_{z}$ is sufficient to model most texts.
    We use uniform distribution as $P_{c}$ because our model is evaluated on single-label datasets in which the most texts are written according to one main topic.
    In a multi-label classification dataset, it is feasible to sample a topic code $c$ containing several topics according to the distribution of human-annotated categories. 
    \item Topic Classifier $Q$\\
    It is a categorical topic code classifier, which predicts the categorical topic distribution from input bag-of-words.
    The categorical classifier $Q$ plays the role of inferring topics from bag-of-words in this work.
    It is trained by predicting the topic code $c$ from the generated bag-of-words $G(c,z)$.
        The  categorical loss $\mathcal{L}_Q$ for training $Q$ is shown as below,
       \begin{equation} \label{eq:L_Q}
   \mathcal{L}_Q =  \E_{z \sim P_{z},c \sim P_{c}, c^\prime=Q( G(z, c))}[ \mathcal{L}_{ce}( c,  c^\prime )  ],
    \end{equation}     
    where the function $\mathcal{L}_{ce}$ evaluates the cross entropy between $c$ and the prediction $c^\prime=Q( G(z, c) )$, and $Q$ learns to minimize the cross entropy. 
    Since $c$ is one-hot vector, cross-entropy is used here. It is possible to use other difference measures.  
    \item Noise predictor $E$\\
    It focuses on predicting the continuous noise $z$ from the bag-of-words $x$ of human-written texts.
    $E$ is only learned with auto-encoder.
\end{itemize}
\subsection{Training} \label{section:training}
We apply (\ref{eq:info_gan}) to train our generator $G$, discriminator $D$, and the topic classifier $Q$.
\begin{equation}\label{eq:info_gan}
    \min_{G, Q} \max_{D} - \mathcal{L}_D + \lambda \mathcal{L}_Q.
\end{equation}
 where $\lambda$ is a hyper-parameter.
As we find that when the value of $\lambda$ is too large, the generator generates unreasonable bag-of-words to minimize $\mathcal{L}_Q$, we set  $\lambda$ to $0.1$.
The discriminator $D$ learns to minimize  $\mathcal{L}_D$,  $Q$ learns to minimize $\mathcal{L}_Q$, and $G$ learns to maximize $\mathcal{L}_D$ while minimizing $\mathcal{L}_Q$.
However, it is difficult to apply infoGAN on textual data.
Hence, we propose some extensions and tips in the following sections.



\subsection{Categorical Loss Clipping}
During training, there is a severe \emph{mode collapse} issue within the same topic. 
That is, given the same topic code $c$, the generator $G$ ignores the continuous noise $z$ and always outputs the same bag-of-words.
The reason is that the optimal solution for the generator to maximize the mutual information (or minimize (\ref{eq:L_Q})) between discrete $c$ and generated discrete bag-of-words is always predicting the same bag-of-words given the same $c$.
 To tackle this issue, we clip the categorical loss in (\ref{eq:L_Q}) to a lower bound $\alpha$ as below.
       \begin{equation} \label{eq:L_Q_prime}
   \mathcal{L}_Q^\prime =  \E_{z \sim P_{z},c \sim P_{c},c^\prime=Q( G(z, c) )}[ max( \mathcal{L}_{ce}( c,c^\prime)  , \alpha ) ].
    \end{equation} 
    With (\ref{eq:L_Q_prime}), the model stops to make the categorical loss smaller when it is small enough, which is controlled by the hyper-parameter $\alpha$. 
    As setting the value of $\alpha$ too large hinders the $Q$ from predicting correct topic code $c$, we set the value slightly greater than zero, which is $0.15$.
It is worth mentioning that applying batch-normalization \cite{batchnorm} to $G$ also greatly alleviates the mode collapse problem.
\subsection{Combining infoGAN with auto-encoder} \label{subsec:ae}
As mentioned by prior works \cite{VAE-GAN,Intro-VAE}, it's useful to train GAN and auto-encoder alternatively.
Therefore, we try to include auto-encoder training in the optimization procedure, as shown in the right part of Figure~\ref{fig:topic_gan}~(A).
This is another way to prevent the mode collapse mentioned in the last subsection. 

We use bag-of-words $x$ from real text to train an auto-encoder.
In auto-encoder training, the noise predictor $E$ and topic classifier $Q$ are jointly regarded as an encoder, which encodes bag-of-words into a continuous code $E(x)$ and a discrete code $Q(x)$ respectively.
Here, the generator $G$ serves as a decoder, which reconstructs the original input $x$ from a continuous code $E(x)$ and a discrete code $Q(x)$.
Therefore,   $G$, $Q$ and $E$   jointly learn to minimize the reconstruction loss, 
\begin{equation}\label{eq:rec_loss}
    \min_{G, Q, E} \E_{x \sim P_{data}, x^\prime=G(Q(x), E(x))}[  \mathcal{L}_{bce}(x , x^\prime   )  ],
\end{equation}
where binary cross entropy loss $\mathcal{L}_{bce}$  is used as the reconstruction loss function.
We train (\ref{eq:info_gan}) and (\ref{eq:rec_loss}) alternately.

\subsection{Using WGAN-gp} 

As mentioned in \cite{WGAN}, it is challenging to generate discrete data using GAN.
Because bag-of-words vectors from real texts are also discrete, when using original loss function for $D$ in  (\ref{eq:L_D}), the training fails.
Here, we apply WGAN \cite{WGAN} loss to train $G$ and $D$ and rewrite (\ref{eq:L_D}) as  
\begin{equation} \label{eq:step1}
\mathcal{L}_D^\prime =   -\E_{x \sim P_{data}}[ D(x) ] 
                    + \E_{z \sim P_{z},c \sim P_{c}}[ D(G(z, c)) ].
\end{equation}
Here, to constrain $D$ to be a Lipschitz function, we apply gradient penalty \cite{WGAN-gp} to $D$.


\subsection{Using pretrained word embeddings}
Prior works \cite{topic-word2vec} suggested that using pretrained word embeddings such as word2vec \cite{word2vec} helps models discover more coherent topics.
To encourage the model to learn more consistent topics, we use pretrained word embeddings on $Q$ to regularize $Q$ to learn more coherent topics.
When $Q$ takes a bag-of-words as input, it first computes the document vector as the weighted average of pretrained word embeddings of bag-of-words. 
Then, it feeds the document vector to a linear network to predict discrete topic code $c$. 
We use smooth inverse frequency (SIF) \cite{tough-to-beat-embedding} to compute the weighted average of the word embeddings.
The weight of word $w$ is $a/(a+p(w))$, where $a$ is a learnable non-negative parameter and $p(w)$ is word frequency.
The word embeddings are pretrained by fasttext~\cite{fasttext}.

\section{Text Generation from Discovered Topics} 
\label{section:text_generation}
To interpret what the model learns, we try to generate text conditioned on discovered discrete topic $c$, which is shown in Figure~\ref{fig:topic_gan}~(B).
Given a discrete topic $c$, we sample many different noises $z$ to generate different bag-of-words $G(c,z)$.
After obtaining bag-of-words from $G$, we use a text generator $G^\prime$ to generate sequential text from the bag-of-words.
Because we can transform each sequential text into its corresponding bag-of-words, we can obtain numerous (bag-of-words, text) pairs to train the text generator $G^\prime$.
However, compared to human-written text, generated $G(c,z)$ may be noisy and unreasonable.
To make $G^\prime$ robust on $G(c,z)$, during training, we add some noise such as word random removal or shuffling to input bag-of-words.
The network architecture of $G^\prime$ is an LSTM whose hidden state is initialized by the output of a feedforward neural network which takes bag-of-words as input.

\section{Experiments} 



\subsection{Unsupervised Classification}
\begin{table*}[t!]
\centering
\begin{tabular}{lcccccc}
\hline
\bf Methods & \bf 20News & \bf Yahoo! & \bf DBpedia &
\bf Stackoverflow & \bf Agnews & \bf News-Cat\\ \hline
word embed avg+k-means & 28.63 & 38.91 & 69.04 & 22.44 & 73.83 & 37.78\\
sent2vec+k-means & 28.06 & 51.24 & 60.98 & 36.78 & 83.82 & 40.64\\
LDA + k-means & 28.97 & 22.58 & 61.13 & 43.57 & 49.36 & 21.84 \\
\hline
NVDM & 24.63 & 33.21 & 46.22 & 26.33 & 62.36 & 30.12 \\
ProdLDA & 30.02 & 39.65 & 68.19 & 25.13 & 72.78 & 38.62 \\
LDA-few topics & 29.62 & 29.44 & 68.62 & 35.82 & 71.07 & 26.24 \\ \hline
KE~\cite{FAC} & 37.8 & \textbf{53.9} & - & - & 73.8 & - \\ \hline
TIGAN & 34.12 & 52.25 & \textbf{85.37} & 47.01 & \textbf{84.13} & \textbf{49.32}\\
TIGAN w/o loss clipping & 30.12 & 45.92 & 83.32 & 40.99 & 83.66 & 47.42\\ 
TIGAN w/o auto-encoder & 29.14 & 43.07 & 78.76 & 31.60 & 81.62 & 45.12\\
TIGAN w/o word embed & 36.89 & 42.14 & 71.26 & 46.14 & 69.78 & 47.32\\
TIGAN w/ linear $Q$ & \textbf{41.01} & 42.91 & 83.73 & \textbf{64.46} & 72.65 & 48.14\\
\hline
\end{tabular}
\caption{Unsupervised classification accuracy.}
\label{table:unsupervised_classification}
\end{table*}

\paragraph{Datasets}
We evaluate our model on 6 various datasets, including (1)20NewsGroups, (2)Yahoo! answers, (3)DBpedia ontology classification, (4)stackoverflow title classification~\cite{stackoverflow} (5)agnews and (6)News-Category-Dataset~\footnote{https://www.kaggle.com/rmisra/news-category-dataset}.

The 20NewsGroups is a document classification dataset composed of 20 different classes of documents.
Yahoo! answers is a question type classification dataset with 10 types of question-answer pairs constructed by \cite{text-from-scratch}.
DBpedia ontology classification dataset is constructed by \cite{text-from-scratch}, with 14 ontology classes selected from DBpedia 2014.
Stackoverflow title classification is constructed by \cite{stackoverflow} with 20 different title categories.
The agnews is a news classification dataset with 4 different categories constructed by \cite{text-from-scratch}.
We combined news titles and descriptions as our training texts.
News-Category-Dataset contains around 200k news headlines from the year 2012 to 2018 obtained from HuffPost.
We selected the most frequent 11 classes and roughly balanced the number of data of each class.



\vspace{-1mm}
\paragraph{Experimental Settings}
For all datasets, we use the same model architecture and same optimizer without tuning on each dataset.
The details of model architecture can be found in Appendix \ref{sec:architecture}.
For each dataset, we choose the most frequent 3000 words as the vocabulary after stop words and punctuation removal and lowercase conversion.
When training TIGAN, we sample a one-hot topic code $c$ from a uniform distribution and set the number of latent topics (i.e. the dimension of $c$) the same as the number of human-annotated categories in each dataset.
For example, as there are four news categories in agnews, we set the topic number to four.
Continuous noise $z$ is sampled from a normal distribution and the dimension of $z$ is set to be 200 in all experiments.
Our model is not sensitive to the dimension of $z$, setting it to smaller value such as $50$ or $100$ will yield similar results.

We use the topic classifier $Q$ to predict the latent topic probability distribution of each sample, and we assign each sample to the latent topic with the maximum probability.
The samples assigned to a latent topic cluster use their human-annotated labels to vote for which label should be assigned to the whole cluster.
After assigning each latent topic to a human-annotated category, we evaluate the classification accuracy as the quality of the captured latent topics.

\vspace{-1mm}
\paragraph{Baselines}
We compare our model to several possible methods for unsupervised text classification, including topic models, unsupervised sentence representation and keywords enrichment (KE) method.

The most straightforward method for unsupervised text classification is unsupervised sentence representation based methods. 
In these methods, we first encode texts into vectors by using unsupervised text representation learning methods and then conduct k-means on the text vectors.
In Table~\ref{table:unsupervised_classification} ``word embed avg+k-means'', we average the pretrained word embeddings as text vectors.
Here, the word embeddings are same as the pretrained word embeddings used in TIGAN.
In Table~\ref{table:unsupervised_classification} ``sent2vec+k-means'', the vectors are extracted by state-of-the-art unsupervised sentence representation method \cite{n-Gram} with latent dimension set to be 300.
In Table~\ref{table:unsupervised_classification} ``LDA + k-means'', we set the topic number of LDA~\cite{blei2003latent} to 50, and then conduct k-means unsupervised clustering on the learned features.

Another baselines are topic models which represent a text as a mixture of latent topics.
Because these latent topics are likely to be identical to human-annotated categories, topic models are possible methods for unsupervised text classification.
The evaluation setting of topic models is same as TIGAN, in which we set the topic number of topic models the same as the number of human-annotated categories and assign each latent topic to its most possible human-annotated category.
For example, in Table~\ref{table:unsupervised_classification} ``LDA-few topic'', differet from the setting of ``LDA + k-means'', the topic number is same as the class number.
Also, two strong variational auto-encoder based neural topic models, including NVDM~\cite{topic-vae} and ProdLDA\cite{ProdLDA} are chosen as our baselines.

Keyword enrichment(KE)~\cite{FAC} is a recent weakly-supervised text classification method which initially asks human experts to label several keywords for each category and then gradually enriches the keywords dictionary.
The documents similar to the keywords of a category are classified to the same category.
This method is not strictly comparable to our method because it requires the help of humans and the performance hinges on good keywords labeled by experts while our model don't require any keyword from human.
Also, it requires WordNet~\cite{wordnet} to retrieve synonym of keywords.

\begin{table*}[t!]
\centering
\begin{tabular}{lcccc}
\hline
\bf Methods & \bf 20NewsGroups & \bf Yahoo! Answers & \bf DBpedia & \bf Gigaword\\ \hline
LDA & 42.68 & 36.34 & 51.06 & 35.61\\
NVDM & 42.36 & 47.02 & 49.95 & 37.22\\
ProdLDA & 43.44  & 52.01 & 55.26 & 43.17 \\
\textbf{TIGAN w/ word embed} & 43.01 & \textbf{55.92} & \textbf{62.16} & \textbf{46.12}\\
\textbf{TIGAN w/o word embed} & \textbf{45.22} & 45.64 & 54.23 & 41.79\\ 
\hline
\end{tabular}
\caption{Topic coherence scores. Higher is better.}
\label{table:coherence}
\end{table*}

\begin{table*}[t!]
\begin{tabularx}{\textwidth}{ccX}
\hline
\bf Dataset & \bf Topic ID & \bf Topical Words \\\hline
\multirow{6}{*}{DBpedia} & 1            
& pianist, composer, singer, cyrillic, songwriter, romania, poet, painter, jazz, actress, musician, tributary \\\cline{2-3}
& 2  & skyscraper, building, courthouse, tower, historic, plaza, brick, floors, twostory, mansion, hotel, buildings, register, tallest, palace \\ \cline{2-3}
& 3 & midfielder, footballer, goalkeeper, football, championship, league, striker, soccer, defender,  goals, matches, cup, hockey, medals\\\hline

\multirow{6}{*}{ English Gigaword} 
& 1  & inflation, index, futures, benchmark, prices, currencies, output, outlook, unemployment, lowest, stocks, opec, mortgage \\ \cline{2-3}
& 2  & sars, environment, pollution, tourism, virus, disease, flights, water, scientific, airports, quality, agricultural, animal, flu, alert \\\cline{2-3}
& 3 & polls, votes, elections, democrats, electoral, election, conservative, re-election, democrat, candidates, republicans, liberal, presidential\\ \hline
\end{tabularx}
\caption{Topical words generated from discovered latent topic code $c$. In DBpedia, Topic 1 is about music, Topic 2 is about building and Topic 3 is about sport. In English Gigaword, Topic 1 is about business, Topic 2 is about disease and Topic 3 is about politics.}
\label{table:topic_words}
\end{table*}

\vspace{-1mm}
\paragraph{Results and Discussion.} 
Compared to the methods of clustering continuous text vectors like ``word embed avg+k-means'' and ``sent2vec+k-means'' in  Table~\ref{table:unsupervised_classification}, TIGAN improves the performance.
The reason is that TIGAN maximize the mutual information between generated data and a discrete one-hot distribution, which encourages classifier $Q$ to learn more salient features to separate texts into different classes.
In ``LDA+k-means'', the result is even worse than ``LDA-few topics'', which means it's not feasible to obtain human-annotated classes by directly clustering the features learned by LDA.

As shown in the Table~\ref{table:unsupervised_classification}, TIGAN outperforms two variational neural topic models and LDA on unsupervised classification.
No matter variational neural topic models or statistical topic models, they all assume  each document is produced from mixture of topics.
To model the variance of documents, they have to split a topic such as ``sport'' into many subtopics such as baseball or basketball.
Because in our unsupervised text classification setting, each document should belong to a single category, this assumption harms the performance.
However, as TIGAN is able to control the variance within a topic with a noise vector, it can directly assume a document is generated from a single main topic, which allows TIGAN to discover the topics identical to human-annotated categories.
This performance gap is understandable because topic models are not originally designed for unsupervised text classification setting.

The performance of TIGAN is on par with keywords enriching (KE) in Table~\ref{table:unsupervised_classification} \footnote{Here, we use the weighted recall reported in the paper because the metric that we compute accuracy score is same as weighted recall.} even without keywords annotated by human experts.
This result suggests that our model is able to automatically categorize texts without any help from human.

\vspace{-1mm}
\subsection{Ablation Study} \label{section:ablation}
\paragraph{Training} We conduct ablation study to show that all mechanisms described in Section~\ref{section:training} are useful.
As shown in Table~\ref{table:unsupervised_classification}, categorical loss clipping improves the performance.
During training, we find clipping this term makes the training more stable because this term is easy to diverge.
Combining infoGAN with auto-encoder also improves the performance.
The possible reason is that it encourages generator to generate realistic data which alleviates the difficulty of discrete data generation.
Although we don't restrict topic code $c$ to be one-hot in auto-encoder training, InfoGAN training forces all the text encoded by $Q$ to be one-hot.

\vspace{-1mm}
\paragraph{Model architecture of Topic Classifier}
In Table~\ref{table:unsupervised_classification}, we find that the model architecture of topic classifier $Q$ greatly influences the unsupervised classification performance.
Both ``w/o word embed'' and ``w/ linear Q'' are the settings without pretrained word embeddings.
We can observe that using pretrained embedding helps topic classifier discover more interpretable topics on most datasets, but not on all datasets.
Similar phenomenon can also be found in Table~\ref{table:coherence}.
As we pretrain word embeddings on each dataset separately, for the datasets that don't contain enough training samples such as stackoverflow dataset or 20NewsGroups dataset, it is difficult to learn representative word embeddings.
Hence, the performance on those datasets is not improved with pretrained word embeddings.
Additionally, we find that using pretrained word embedding makes $Q$ not sensitive to the random initialization weights, and thus $Q$ discovers more consistent topics at each training.

In ``w/o word embed'', the model architecture of $Q$ is same as TIGAN (i.e. a neural network with randomly initialized word embeddings), while in ``w/ linear Q'', the model model architecture of $Q$ is simply a linear network without any word embeddings.
Comparing ``w/o word embed'' with ``w/ linear Q'', the simpler model architecture yields better results.
With more complex model of $Q$, it arbitrarily maps complex but not meaningful distribution to one-hot code, which makes the learned topics not interpretable.

\begin{table*}[t!]
\begin{tabularx}{\textwidth}{ccX}
\hline
\bf Dataset & \bf Topic ID        & \bf Generated Text \\\hline
\multirow{6}{*}{DBpedia} & 1 &  james nelson is an american musician singer songwriter and actor he was a line of his work with musical career as metal albums in 1989 \\\cline{2-3}
& 2  & the UNK is a skyscraper located in downtown washington dc district it was completed june 2009 and tallest building currently  \\ \cline{2-3}
& 3 & UNK born 3 january 1977 is a finnish football player who plays for west coast club dynamo he has won bronze medals at 2007 \\\hline
\multirow{6}{*}{English Gigaword}
& 1  &  oil prices fell in asia friday as traders fear of the world demand for biggest drop summer months figures \\ \cline{2-3}
& 2  &  the world health organisation has placed on bird flu in nigeria 's most populous countries virus , saying they are expected to appear \\\cline{2-3}
& 3 & the ruling party won overwhelming majority of seats in opposition 's battle for sweeping elections \\ \hline
\end{tabularx}
\caption{Texts generated from topics with topic ID corresponding to Table~\ref{table:topic_words}. 
The texts in the same dataset are generated from different $c$ but from the same continuous noise $z$.}
\label{table:topic_text}
\end{table*}
\begin{table*}[t!]
\begin{tabularx}{\textwidth}{X}
\hline
\bf Generated Text \\\hline
 UNK is a historic public square located in south bronx built in 1987 \\ \hline
 UNK is a french southwest hotel originally located in the of greece south carolina it was conceived by josef  \\ \hline
 UNK in historic district is a methodist church located in south african cape was added to national register of places 200 \\\hline
\end{tabularx}
\caption{The generated texts of DBpedia Topic ID 2 in Table~\ref{table:topic_words} with different noise vectors $z$.}
\label{table:same_topic_text}
\end{table*}

\vspace{-2mm}
\subsection{Topic Coherence}
\vspace{-1mm}
In this section, we use the method in Appendix~\ref{sec:retrieve_topical_words} to extract topical words to represent latent topics and evaluate the quality of the topical words.
As topic models are also good at extracting topical words to represent topics, they are chosen as our baselines.

\vspace{-1mm}
\paragraph{Experimental setup}
In this section, the topic number of all models is same as the number of human-annotated classes.
To show our model not only works on classification datasets, we try to evaluate our model on an unannotated dataset which is English Gigaword~\cite{abs-summarization}, a news summarization dataset.
The topic number on English Gigaword is set to be 10 and the discussion of how to decide topic code distribution is in Appendix~\ref{sec:P_c}.

\vspace{-1mm}
\paragraph{Quantitative analysis.}
We use the $C_{v}$ metric \cite{CV_metric} to evaluate the coherence of extracted topical words~\cite{interpret-topic}. \footnote{We use the CoherenceModel in gensim module to compute coherence scores.}
The coherence scores are computed by using English Wikipedia of 5.6 million articles as an external corpus.
The higher coherence scores in Table \ref{table:coherence} show that TIGAN extracts reasonable and coherent words to represent a topic.

\paragraph{Qualitative analysis.}
To further analyze the quality of discovered topics, the topical words of latent topics are listed in Table~\ref{table:topic_words}.
Additionally, we use the method mentioned in Section~
\ref{section:text_generation} to generate corresponding texts of the discovered topics in Table~\ref{table:topic_text}.
In the tables, the captured latent topics are clearly semantically different based on the generated topical words.
Similarly, the generated texts are also topically related to the associated topics.
More extracted topical words and generated texts are offered in Appendix~\ref{sec:ana_qualitative}.


\subsection{{Disentangled Representations}}

The essential assumption of our work is that $c$ and $z$ can learn disentangled representations, where $c$ encodes main topic information and $z$ control the variance within the topics.
To validate this assumption, we generate texts from the same continuous noise $z$ but from different topic code $c$ in Table~\ref{table:topic_text}.
There is almost no overlap of words between the texts generated from the same continuous noise $z$ but from different $c$, which means $z$ encodes the information disentangled from the topic code $c$. 

We also want to analyse whether $z$ can control the variance within the topic by modeling subtopic information.
In Table~\ref{table:same_topic_text} and Appendix.Table~\ref{table:more_same_topic_text}, we generate the texts from the same topic code $c$ but from different noise $z$.
It is obvious that in Table~\ref{table:same_topic_text} the topic of $c$ is about \textit{building}, but with different $z$ quite different sentences are generated.
The diversity of generated texts within the same topic shows $z$ is capable to model the variance within the same topic.
Also, by generating numerous texts of a same topic, it gives us a way to interpret the discovered topics further.
With the above two experiments, we demonstrate that our model is able to learn disentangled representations. 

\vspace{-1mm}
\section{Conclusion}
\vspace{-1mm}
This paper proposes a novel unsupervised framework, TIGAN, for exploring unsupervised discrete text representations to interpret textual data.
By learning a discrete topic code disentangled from a continuous vector controlling subtopic information, TIGAN shows the superior performance on unsupervised text classification, which gives humans a good way to understand textual data.
The extracted topical words and generated texts from the discovered topics further showcase the effectiveness of our model to learn explainable and coherent topics.

\bibliography{anthology,acl2020}
\bibliographystyle{acl_natbib}

\appendix
\clearpage

\section{Model Architecture and Hyper-parameters} \label{sec:architecture}
In all datasets, we use the same model architecture and same optimizer without tuning our model on each dataset.
The model architecture of generator $G$ is a fully-connected neural network with 3 hidden layers and the hidden dimension is 1000 of each hidden layer.
Setting hidden dimension to 500 or 1500 only slightly influences the performance.
We apply batch normalization~\cite{batchnorm} at each layer.
The discriminator $D$ is a  fully-connected neural network with 2 hidden layer and the hidden dimension is 500 of each hidden layer.
As the discriminator has to be well trained to guide the gradient of generator, we make discriminator easier to train that the model architecture of discriminator is simpler than generator.
If the hidden dimension of $G$ and $D$ is too deep, $G$ and $D$ becomes difficult to train.
On the other hand, if the hidden dimension of $G$ and $D$ is too shallow, the performance of TIGAN drops.

The topic classifier $Q$ is consisted of a word embedding matrix and a linear network which takes the weighted average of word embeddings as input and predicts the latent topics.
We set the dimension of pretrained word embedding to 100.
As we find the model architecture of topic classifier $Q$ greatly influences the performance of our model, we has discussed the model architecture of $Q$ in later section.
The noise predictor $E$ is a fully-connected neural network with one hidden layer of dimension 500.
The $G$,$D$, $E$ and $Q$ are all optimized by Adam optimizer with learning rate $0.0005$, $\beta_{1}=0.5$ and $\beta_{2}=0.999$.

\section{Retrieving Topical Words from TIGAN} \label{sec:retrieve_topical_words}
The topical words of our model can be retrieved from topic classifier $Q$ as following.
$Q$ consists of a word embedding matrix and a linear layer.
Let $W_{V \times N}$ be a word embedding matrix, where each row is a $N$ dimensional word vector, and $V$ is the vocabulary size.
Let $M_{K \times N}$ be the weight matrix of the linear layer to predict the topics from sentence embedding which is the weighted average of word embeddings, where $K$ is the topic number.
Let $C_{K \times V}=M \cdot W^{T}$, where the value of $C_{k,v}$ represents the importance of the v-th word to the k-th topic.
The top few words with the highest values within row $k$ are selected as the topical words of k-th topic.

\section{Qualitative Analysis of Topical Words} \label{sec:ana_qualitative}
We provide the learned topical words of all latent topics of Yahoo! Answer from our model in Table~\ref{table:yahoo_topic_words_TIGAN}, ProdLDA in Table~\ref{table:yahoo_topic_words_ProdLDA}, NVDM in Table~\ref{table:yahoo_topic_words_nvdm} and LDA in Table~\ref{table:yahoo_topic_words_LDA}.
As shown in the tables, the topical words in each latent topic of TIGAN are closely semantically related to each other.
For example, all the words in the first latent topic are obviously related to science, the words in second topic are related to the operation of computer and the words in third topic are related to politics.

Most of the topical words from ProdLDA are all related to each other, except some words.
For example, some words in the second-last latent topic is not reasonable like ``cousin'', ``do I'' or ``bc''.
In NVDM, we can find more unreasonable latent topics.
For example, in the last latent topic, ``diet'' and ``movie'' should not be clustered into the same topic and in fourth-last latent topic, ``immigrants'' and ``cup'' is not related to each other.
The quality of extracted topical words from LDA is worse than the above three neural models.
Most of the topical words in a same topic are not semantically related to each other. 

\section{Discussion of Topic Code Distribution} \label{sec:P_c}
For labeled datasets, no matter single-label or multi-label datasets, we encourage to sample topic code  $c$ according to the distribution of the human-labeled categories.
For unlabeled datasets, we encourage to set the topic number of $c$ to $10 \sim 20$ and tune from greater number.
With a greater topic number, TIGAN is still able to learn explainable topics as this setting just splits a single topic into several subtopics.
However, setting topic number smaller forces several unrelated topics to combine into a single topic and thus harms the performance.
It's worth mentioning that we find that sampling one-hot $c$ from uniform distribution works for most datasets.
That is because it encourages topic classifier to learn salient and highly interpretable features to maximize the mutual information between discrete one-hot $c$ and generated discrete bag-of-words.

\begin{table*}[t!]
\begin{tabularx}{\textwidth}{X}
\hline
 \bf Generated Text \\\hline
 british airports leaders plan to ban on one of thousands the city health virus officials said friday \\\hline
 country health officials from world organisation has warned turkey 's swine flu in britain pakistan saying it was unsafe bringing the bird \\\hline
 the talk of a new disease on tiny team of has formed in tanzania which based on ebola virus \\\hline
 sars epidemic has been recently detected in the hong kong and disease in france public health organisation is for \\\hline
 health experts said it was in southwest china to contain the outbreak of disease \\\hline
 national organizations have been established in beijing to sars for safety alert and common  disease resources \\\hline
 the swine flu virus has been muted by health organisation response to tackle disease \\\hline
 thailand 's sichuan province has given to foreign investors in airlines travel disease \\\hline
 indonesia has developed a discovery of health system wild birds at home from the flu \\\hline
 the impact of poor health organisation has been UNK swine flu season in for population \\\hline
 the governments gathered in thailand 's health officials of world ministry earlier on friday \\\hline
 indonesia has been hospitalized with pneumonia the health organisation said \\\hline
doctors from romania and countries are to explore ways of recent disease \\\hline
the death of \# percent bird flu in latest report region asia , according to a study released by u.s. health organization \\\hline
agricultural experts have set to double the european commission meet with a new study on and human bird flu in animal was unveiled \\\hline
\end{tabularx}
\caption{The generated texts of English Gigaword Topic ID 2 in Table~\ref{table:topic_words} with different noise vectors. It's clear that all the generated sentences are about disease.}
\label{table:more_same_topic_text}
\end{table*}

\begin{table*}[t!]
\begin{tabularx}{\textwidth}{X}
\hline
 \bf Topical Words \\\hline
  measure, equations, cm, density, equation, units, formula, solar, atoms, inches, cycle, calculate, volume, radius, physics, triangle, molecules, height, equal, gravity
  \\\hline
  deleted, files, folder, users, delete, messenger, spyware, scan, installed, settings, upload, downloaded, ip, disk, router, screen, firewall, software, message, xp
  \\ \hline
  terrorists, democrats, troops, republicans, terrorism, terrorist, leaders, senate, iraq, politicians, liberals, democracy, saddam, congress, democratic, elections, elected, political, majority, war \\ \hline
  biology, engineering, courses, resources, management, materials, algebra, technology, design, accounting, analysis, studies, communication, learning, education, programming, skills, teaching, study, marketing
  \\\hline
  nervous, abuse, depressed, drunk, crying, depression, orgasm, jealous, sexually, jail, hang, drinking, emotional, chat, busy, sleeping, herself, hanging, custody, diagnosed \\\hline
  sore, infection, skin, stomach, urine, milk, vitamin, pills, symptoms, muscle, therapy, breast, severe, treatment, muscles, fluid, bleeding, acne, pain \\\hline 
  bone spirit, holy, heaven, rock, satan, worship, ghost, evolution, spiritual, bible, devil, jesus, bands, mary, lyrics, adam, gods, soul, singer, christian \\\hline
  chat, pics, porn, orgasm, females, nasty, bored, penis, sensitive, avatar, attracted, emotional, ugly, naked, sexy, addicted, vagina, lesbian, jokes, dirty \\\hline
  champions, playoffs, championship, league, teams, cup, nba, team, finals, nfl, hockey, fifa, football, soccer, kobe, wrestling, basketball, brazil, matches, cricket \\\hline
  payments, estate, mortgage, property, loan, insurance, payment, fees, loans, funds, debt, companies, investment, agency, financial, employees, filed, employment, owner, taxes \\\hline
\end{tabularx}
\caption{Topical words generated from TIGAN in Yahoo! Answers.}
\label{table:yahoo_topic_words_TIGAN}
\end{table*}

\begin{table*}[t!]
\begin{tabularx}{\textwidth}{X}
\hline
 \bf Topical Words \\\hline
    teaching ,learning ,subject ,teach ,learn ,skills ,studying ,topic ,study ,project ,prepare ,courses ,students ,schools ,english ,guide ,essay ,language ,spelling ,education \\\hline
    tooth ,teeth ,knee ,dentist ,exercise ,loose ,dr ,healthy ,coffee ,gym ,counter ,workout ,stomach ,pill ,medication ,severe ,acne ,bleeding ,eating ,diet \\\hline
    fees ,employees ,employee ,employer ,cash ,agent ,earn ,selling ,homes ,banks ,hire ,mortgage ,payment ,invest ,offered ,funds ,budget ,estate ,bills ,agency \\\hline
    divide ,motion ,factor ,formula ,direction ,scale ,compound ,length ,equations ,circle ,zero ,frequency ,steel ,wave ,triangle ,elements ,electricity ,velocity ,solid ,table \\\hline
    detroit ,olympics ,nfl ,league ,winning ,wins ,teams ,2002 ,playoffs ,match ,championship ,johnson ,coach ,bowl ,dallas ,team ,goal ,soccer ,baseball ,finals \\\hline
    liberals ,congress ,conservative ,voting ,911 ,saddam ,liberal ,soldiers ,majority ,elected ,bush ,politicians ,vote ,elections ,violence ,terrorism ,freedom ,vietnam ,troops ,democracy \\\hline
    band ,bob ,bands ,lyrics ,singing ,artist ,singer ,scene ,sang ,80s ,movie ,episode ,sings ,sing ,pink ,rap ,dancing ,movies ,song ,songs \\\hline
    bible ,believe ,gods ,christianity ,christ ,worship ,faith ,spiritual ,jesus ,spirit ,belief ,christian ,god ,heaven ,jewish ,christians ,eve ,mary ,lord ,belive \\\hline
    dated ,talked ,weve ,eachother ,cousin ,friendship ,ex ,gf ,break ,cheating ,cheated ,hang ,dating ,together ,broke ,boyfriends ,bc ,hanging ,doi ,talk\\\hline
    opened ,icon ,automatically ,task ,edition ,comp ,blocked ,sharing ,instant ,deleted ,press ,blank ,dell ,keyboard ,connection ,bar ,beta ,update ,mac ,dsl\\\hline

\end{tabularx}
\caption{Topical words generated from ProdLDA in Yahoo! Answers.}
\label{table:yahoo_topic_words_ProdLDA}
\end{table*}

\begin{table*}[t!]
\begin{tabularx}{\textwidth}{X}
\hline
 \bf Topical Words \\\hline
   earn ,points ,money ,countries ,questions ,energy ,learn ,invest ,score ,market ,nuclear ,iran ,dollars ,business ,technology ,government ,study ,stock ,economy ,world\\\hline
    school ,schools ,high ,university ,college ,colleges ,students ,basketball ,sport ,la ,de ,girls ,grade ,courses ,graduate ,teacher ,football ,les ,teachers ,boys\\\hline
    english ,lyrics ,song ,search ,google ,spanish ,translate ,site ,language ,christian ,sites ,myspace ,sings ,bible ,love ,translation ,websites ,songs ,web ,books\\\hline
    god ,jesus ,christians ,christian ,religion ,truth ,bush ,religious ,beliefs ,faith ,church ,christ ,believe ,evil ,heaven ,feelings ,bible ,kill ,gods ,illegal\\\hline
    download ,computer ,software ,email ,myspace ,laptop ,install ,phone ,pc ,card ,cd ,program ,online ,password ,files ,video ,address ,send ,connect ,downloaded\\\hline
    relationship ,sex ,math ,grade ,age ,degree ,feelings ,shy ,boyfriend ,teacher ,homework ,advice ,healthy ,pregnant ,study ,dating ,bf ,classes ,yrs ,weight\\\hline
    win ,hate ,watch ,immigrants ,idol ,americans ,mexicans ,hes ,gonna ,watching ,iraq ,games ,tired ,mexico ,jobs ,thinks ,illegals ,cup ,republicans ,democrats\\\hline
    foods ,food ,health ,diet ,products ,skin ,protein ,fat ,muscle ,eat ,eating ,exercise ,treatment ,cure ,disease ,healthy ,muscles ,sugar ,bacteria ,weight\\\hline
    county ,tax ,state ,federal ,income ,insurance ,loan ,taxes ,visa ,citizen ,pay ,financial ,states ,department ,lawyer ,child ,california ,legal ,loans ,court\\\hline
    christmas ,season ,diet ,dvd ,burn ,buy ,songs ,movie ,night ,water ,song ,day ,band ,summer ,drink ,gift ,eat ,ebay ,credit ,price\\\hline
\end{tabularx}
\caption{Topical words generated from NVDM in Yahoo! Answers.}
\label{table:yahoo_topic_words_nvdm}
\end{table*}

\begin{table*}[t!]
\begin{tabularx}{\textwidth}{X}
\hline
 \bf Topical Words \\\hline
looking, friend, girl, yahoo, song, new, look, good, trying, know, im, information, music, online, business, need, write, check, send, change \\\hline
like, know, want, really, think, people, help, say, question, tell, answer, need, good, make, guy, feel, things, thing, bad, ask \\\hline
school, job, work, home, state, high, good, college, number, age, small, book, run, law, area, want, best, parents, doctor, vs \\\hline
time, best, long, person, years, money, man, life, sex, guys, like, women, email, way, great, little, lot, men, away, good \\\hline
day, real, read, bush, anybody, eat, explain, 12, power, pain, mind, food, new, red, period, rid, weeks, air, hours, ex \\\hline
water, times, end, english, card, point, states, difference, credit, page, gets, comes, different, 10, blood, light, time, used, order, green \\\hline
need, help, free, site, computer, want, website, know, heard, try, weight, using, web, best, whats, download, windows, like, company, search \\\hline
people, god, called, believe, country, think, pay, days, ago, means, war, child, come, world, probably, president, info, wondering, usa, test \\\hline
use, old, better, internet, told, buy, type, stop, used, black, white, ones, turn, easy, language, date, science, fast, laptop, skin \\\hline
love, going, think, got, world, year, talk, yes, team, body, win, hes, come, american, movie, game, play, cup, ive, boyfriend \\\hline

\end{tabularx}
\caption{Topical words generated from LDA in Yahoo! Answers.}
\label{table:yahoo_topic_words_LDA}
\end{table*}

\end{document}